\documentclass[11pt,letterpaper]{article}
\usepackage{emnlp2016}
\usepackage{pslatex}
\usepackage{latexsym}
\usepackage{tikz}
\usepackage{tikz-dependency}
\usepackage{float}
\usepackage{latexsym}  
\usepackage{url}
\usepackage{amsmath}
\usepackage{verbatim}
\usepackage{graphicx,adjustbox}
\usepackage{subfig}
\usepackage{pslatex}
\usepackage{tikz}
\usepackage{xcolor}
\usepackage{array}
\newcolumntype{L}[1]{>{\raggedright\let\newline\\\arraybackslash\hspace{0pt}}m{#1}}
\newcolumntype{C}[1]{>{\centering\let\newline\\\arraybackslash\hspace{0pt}}m{#1}}
\newcolumntype{R}[1]{>{\raggedleft\let\newline\\\arraybackslash\hspace{0pt}}m{#1}}
\usetikzlibrary{fadings}

\def\emnlppaperid{***}

\emnlpfinalcopy

\title{Long Short-Term Memory-Networks for Machine Reading}

\author{Jianpeng Cheng, Li Dong \and Mirella Lapata\\
School of Informatics, University of Edinburgh\\
10 Crichton Street, Edinburgh EH8 9AB\\
\texttt{\{jianpeng.cheng,li.dong\}@ed.ac.uk}, \texttt{mlap@inf.ed.ac.uk}
}

\begin{document}
	\maketitle
	\begin{abstract}

          In this paper we address the question of how to render
          sequence-level networks better at handling structured input.
          We propose a machine reading simulator which processes text
          incrementally from left to right and performs shallow
          reasoning with memory and attention. The reader extends the
          Long Short-Term Memory architecture with a memory network in
          place of a single memory cell. This enables adaptive memory
          usage during recurrence with neural attention, offering a
          way to weakly induce relations among tokens.  The system is
          initially designed to process a single sequence but we also
          demonstrate how to integrate it with an encoder-decoder
          architecture.  Experiments on language modeling, sentiment
          analysis, and natural language inference show that our model
          matches or outperforms the state of the art.
		
	\end{abstract}
	

	\section{Introduction}

        How can a sequence-level network induce relations which are
        presumed latent during text processing? How can a recurrent
        network attentively memorize longer sequences in a way that
        humans do?  In this paper we design a machine reader that
        automatically learns to understand text.  The term machine
        reading is related to a wide range of tasks from answering
        reading comprehension questions \cite{Clark:ea:2013}, to fact
        and relation extraction
        \cite{Etzioni:etal:2011,fader-soderland-etzioni:2011:EMNLP},
        ontology learning \cite{poon-domingos:2010:ACL}, and textual
        entailment \cite{Dagan:ea:2005}. Rather than focusing on a
        specific task, we develop a general-purpose reading simulator,
        drawing inspiration from human language processing and the
        fact language comprehension is incremental with readers
        continuously extracting the meaning of utterances on a
        word-by-word basis.


	
	In order to understand texts, our machine reader should
        provide facilities for extracting and representing meaning
        from natural language text, storing meanings internally, and
        working with stored meanings to derive further consequences.
        Ideally, such a system should be robust, open-domain, and
        degrade gracefully in the presence of semantic representations
        which may be incomplete, inaccurate, or incomprehensible.  It
        would also be desirable to simulate the behavior of English
        speakers who process text sequentially, from left to right,
        fixating nearly every word while they read
        \cite{rayner2009eye} and creating partial representations for
        sentence prefixes \cite{Konieczny:2000,Tanenhaus:ea:1995}.

        Language modeling tools such as recurrent neural networks
        (RNN) bode well with human reading behavior
        \cite{Frank:Bod:2011}.  RNNs treat each sentence as a sequence
        of words and recursively compose each word with its previous
        \textit{memory}, until the meaning of the whole sentence has
        been derived.  In practice, however, sequence-level networks
        are met with at least three challenges. The first one concerns
        model training problems associated with vanishing and
        exploding gradients
        \cite{hochreiter1991untersuchungen,bengio1994learning}, which
        can be partially ameliorated with gated activation functions,
        such as the Long Short-Term Memory (LSTM)
        \cite{hochreiter1997long}, and gradient clipping
        \cite{pascanu2012difficulty}. The second issue relates to
        memory compression problems. As the input sequence gets
        compressed and blended into a single dense vector,
        sufficiently large memory capacity is required to store past
        information. As a result, the network generalizes poorly to
        long sequences while wasting memory on shorter ones. Finally,
        it should be acknowledged that sequence-level networks lack a
        mechanism for handling the structure of the input. This
        imposes an inductive bias which is at odds with the fact that
        language has inherent structure. In this paper, we develop a
        text processing system which addresses these limitations while
        maintaining the incremental, generative property of a
        recurrent language model.

	\begin{figure}[t]
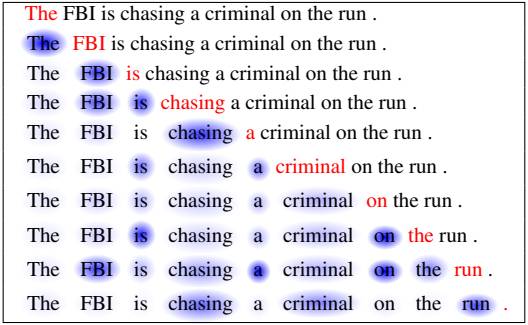

		\centering
		\scriptsize
		\begin{tabular}{|l|}
			\hline 
			{  \color{red}The} FBI is chasing a criminal on the run . \\
			
			\tikz[baseline=(A.base)]{\node[opacity=-1](A) {The};
				\shade[inner color=blue!100] (A.south east) rectangle (A.north west);
				\path (A.center)
                  \pgfextra{\pgftext{The}};} {\color{red}FBI} is
                  chasing a criminal on the run . \\
			
			\tikz[baseline=(A.base)]{\node[opacity=0](A) {The};
				\shade[inner color=blue!10] (A.south east) rectangle (A.north west);
				\path (A.center) \pgfextra{\pgftext{The}};} \tikz[baseline=(A.base)]{\node[opacity=0](A) {FBI};
				\shade[inner color=blue!80] (A.south east) rectangle (A.north west);
				\path (A.center)
                  \pgfextra{\pgftext{FBI}};}  {\color{red}is} chasing
                  a criminal on the run . \\
			
			\tikz[baseline=(A.base)]{\node[opacity=0](A) {The};
				\shade[inner color=blue!15] (A.south east) rectangle (A.north west);
				\path (A.center) \pgfextra{\pgftext{The}};} \tikz[baseline=(A.base)]{\node[opacity=0](A) {FBI};
				\shade[inner color=blue!70] (A.south east) rectangle (A.north west);
				\path (A.center) \pgfextra{\pgftext{FBI}};} \tikz[baseline=(A.base)]{\node[opacity=0](A) {is};
				\shade[inner color=blue!80] (A.south east) rectangle (A.north west);
				\path (A.center)
                  \pgfextra{\pgftext{is}};} {\color{red}chasing} a
                  criminal on the run . \\
			
			\tikz[baseline=(A.base)]{\node[opacity=0](A) {The};
				\shade[inner color=blue!10] (A.south east) rectangle (A.north west);
				\path (A.center) \pgfextra{\pgftext{The}};} \tikz[baseline=(A.base)]{\node[opacity=0](A) {FBI};
				\shade[inner color=blue!20] (A.south east) rectangle (A.north west);
				\path (A.center) \pgfextra{\pgftext{FBI}};} \tikz[baseline=(A.base)]{\node[opacity=0](A) {is};
				\shade[inner color=blue!0] (A.south east) rectangle (A.north west);
				\path (A.center) \pgfextra{\pgftext{is}};} \tikz[baseline=(A.base)]{\node[opacity=0](A) {chasing};
				\shade[inner color=blue!80] (A.south east) rectangle (A.north west);
				\path (A.center)
                  \pgfextra{\pgftext{chasing}};} {\color{red}a}
                  criminal on the run . \\
			
			\tikz[baseline=(A.base)]{\node[opacity=0](A) {The};
				\shade[inner color=blue!10] (A.south east) rectangle (A.north west);
				\path (A.center) \pgfextra{\pgftext{The}};} \tikz[baseline=(A.base)]{\node[opacity=0](A) {FBI};
				\shade[inner color=blue!10] (A.south east) rectangle (A.north west);
				\path (A.center) \pgfextra{\pgftext{FBI}};} \tikz[baseline=(A.base)]{\node[opacity=0](A) {is};
				\shade[inner color=blue!60] (A.south east) rectangle (A.north west);
				\path (A.center) \pgfextra{\pgftext{is}};} \tikz[baseline=(A.base)]{\node[opacity=0](A) {chasing};
				\shade[inner color=blue!20] (A.south east) rectangle (A.north west);
				\path (A.center) \pgfextra{\pgftext{chasing}};} \tikz[baseline=(A.base)]{\node[opacity=0](A) {a};
				\shade[inner color=blue!60] (A.south east) rectangle (A.north west);
				\path (A.center) \pgfextra{\pgftext{a}};} {\color{red}criminal} on the run . \\
			
			\tikz[baseline=(A.base)]{\node[opacity=0](A) {The};
				\shade[inner color=blue!10] (A.south east) rectangle (A.north west);
				\path (A.center) \pgfextra{\pgftext{The}};} \tikz[baseline=(A.base)]{\node[opacity=0](A) {FBI};
				\shade[inner color=blue!20] (A.south east) rectangle (A.north west);
				\path (A.center) \pgfextra{\pgftext{FBI}};} \tikz[baseline=(A.base)]{\node[opacity=0](A) {is};
				\shade[inner color=blue!30] (A.south east) rectangle (A.north west);
				\path (A.center) \pgfextra{\pgftext{is}};} \tikz[baseline=(A.base)]{\node[opacity=0](A) {chasing};
				\shade[inner color=blue!20] (A.south east) rectangle (A.north west);
				\path (A.center) \pgfextra{\pgftext{chasing}};} \tikz[baseline=(A.base)]{\node[opacity=0](A) {a};
				\shade[inner color=blue!30] (A.south east) rectangle (A.north west);
				\path (A.center) \pgfextra{\pgftext{a}};} \tikz[baseline=(A.base)]{\node[opacity=0](A) {criminal};
				\shade[inner color=blue!30] (A.south east) rectangle (A.north west);
				\path (A.center) \pgfextra{\pgftext{criminal}};} {\color{red}on} the run . \\
			
			\tikz[baseline=(A.base)]{\node[opacity=0](A) {The};
				\shade[inner color=blue!10] (A.south east) rectangle (A.north west);
				\path (A.center) \pgfextra{\pgftext{The}};} \tikz[baseline=(A.base)]{\node[opacity=0](A) {FBI};
				\shade[inner color=blue!20] (A.south east) rectangle (A.north west);
				\path (A.center) \pgfextra{\pgftext{FBI}};} \tikz[baseline=(A.base)]{\node[opacity=0](A) {is};
				\shade[inner color=blue!90] (A.south east) rectangle (A.north west);
				\path (A.center) \pgfextra{\pgftext{is}};} \tikz[baseline=(A.base)]{\node[opacity=0](A) {chasing};
				\shade[inner color=blue!20] (A.south east) rectangle (A.north west);
				\path (A.center) \pgfextra{\pgftext{chasing}};} \tikz[baseline=(A.base)]{\node[opacity=0](A) {a};
				\shade[inner color=blue!30] (A.south east) rectangle (A.north west);
				\path (A.center) \pgfextra{\pgftext{a}};} \tikz[baseline=(A.base)]{\node[opacity=0](A) {criminal};
				\shade[inner color=blue!30] (A.south east) rectangle (A.north west);
				\path (A.center) \pgfextra{\pgftext{criminal}};} \tikz[baseline=(A.base)]{\node[opacity=0](A) {on};
				\shade[inner color=blue!100] (A.south east) rectangle (A.north west);
				\path (A.center) \pgfextra{\pgftext{on}};} {\color{red}the} run . \\
			
			\tikz[baseline=(A.base)]{\node[opacity=0](A) {The};
				\shade[inner color=blue!10] (A.south east) rectangle (A.north west);
				\path (A.center) \pgfextra{\pgftext{The}};} \tikz[baseline=(A.base)]{\node[opacity=0](A) {FBI};
				\shade[inner color=blue!80] (A.south east) rectangle (A.north west);
				\path (A.center) \pgfextra{\pgftext{FBI}};} \tikz[baseline=(A.base)]{\node[opacity=0](A) {is};
				\shade[inner color=blue!30] (A.south east) rectangle (A.north west);
				\path (A.center) \pgfextra{\pgftext{is}};} \tikz[baseline=(A.base)]{\node[opacity=0](A) {chasing};
				\shade[inner color=blue!40] (A.south east) rectangle (A.north west);
				\path (A.center) \pgfextra{\pgftext{chasing}};} \tikz[baseline=(A.base)]{\node[opacity=0](A) {a};
				\shade[inner color=blue!100] (A.south east) rectangle (A.north west);
				\path (A.center) \pgfextra{\pgftext{a}};} \tikz[baseline=(A.base)]{\node[opacity=0](A) {criminal};
				\shade[inner color=blue!30] (A.south east) rectangle (A.north west);
				\path (A.center) \pgfextra{\pgftext{criminal}};} \tikz[baseline=(A.base)]{\node[opacity=0](A) {on};
				\shade[inner color=blue!100] (A.south east) rectangle (A.north west);
				\path (A.center) \pgfextra{\pgftext{on}};} \tikz[baseline=(A.base)]{\node[opacity=0](A) {the};
				\shade[inner color=blue!50] (A.south east) rectangle (A.north west);
				\path (A.center) \pgfextra{\pgftext{the}};} {\color{red}run} . \\
			
			\tikz[baseline=(A.base)]{\node[opacity=0](A) {The};
				\shade[inner color=blue!10] (A.south east) rectangle (A.north west);
				\path (A.center) \pgfextra{\pgftext{The}};} \tikz[baseline=(A.base)]{\node[opacity=0](A) {FBI};
				\shade[inner color=blue!10] (A.south east) rectangle (A.north west);
				\path (A.center) \pgfextra{\pgftext{FBI}};} \tikz[baseline=(A.base)]{\node[opacity=0](A) {is};
				\shade[inner color=blue!5] (A.south east) rectangle (A.north west);
				\path (A.center) \pgfextra{\pgftext{is}};} \tikz[baseline=(A.base)]{\node[opacity=0](A) {chasing};
				\shade[inner color=blue!60] (A.south east) rectangle (A.north west);
				\path (A.center) \pgfextra{\pgftext{chasing}};} \tikz[baseline=(A.base)]{\node[opacity=0](A) {a};
				\shade[inner color=blue!10] (A.south east) rectangle (A.north west);
				\path (A.center) \pgfextra{\pgftext{a}};} \tikz[baseline=(A.base)]{\node[opacity=0](A) {criminal};
				\shade[inner color=blue!50] (A.south east) rectangle (A.north west);
				\path (A.center) \pgfextra{\pgftext{criminal}};} \tikz[baseline=(A.base)]{\node[opacity=0](A) {on};
				\shade[inner color=blue!10] (A.south east) rectangle (A.north west);
				\path (A.center) \pgfextra{\pgftext{on}};} \tikz[baseline=(A.base)]{\node[opacity=0](A) {the};
				\shade[inner color=blue!5] (A.south east) rectangle (A.north west);
				\path (A.center) \pgfextra{\pgftext{the}};} \tikz[baseline=(A.base)]{\node[opacity=0](A) {run};
				\shade[inner color=blue!75] (A.south east) rectangle (A.north west);
				\path (A.center) \pgfextra{\pgftext{run}};} {\color{red}.} \\
			\hline
		\end{tabular}
		\caption{Illustration of our model while reading the
                  sentence \textsl{The FBI is chasing a criminal on
                    the run}. Color \textit{red} represents the
                  current word being fixated, \textit{blue} represents
                  memories. Shading indicates the degree of memory
                  activation.}
		\label{example}
		\vspace{-3ex}
	\end{figure}

        Recent attempts to render neural networks more structure aware
        have seen the incorporation of external memories in the
        context of recurrent neural networks
        \cite{weston2014memory,sukhbaatar2015end,grefenstette2015learning}. The
        idea is to use multiple memory slots outside the recurrence to
        piece-wise store representations of the input; read and write
        operations for each slot can be modeled as an attention
        mechanism with a recurrent controller.  We also leverage
        memory and attention to empower a recurrent network with
        stronger memorization capability and more importantly the
        ability to discover relations among tokens.  This is realized
        by inserting a memory network module in the update of a
        recurrent network together with attention for memory
        addressing.  The attention acts as a weak inductive module
        discovering relations between input tokens, and is trained
        without direct supervision.  As a point of departure from
        previous work, the memory network we employ is internal to the
        recurrence, thus strengthening the interaction of the two and
        leading to a representation learner which is able to reason
        over shallow structures.  The resulting model, which we term
        Long Short-Term Memory-Network (LSTMN), is a reading simulator
        that can be used for sequence processing tasks.



	Figure~\ref{example} illustrates the reading behavior of the
        LSTMN.  The model processes text incrementally while learning
        which past tokens in the memory and to what extent they relate
        to the current token being processed. As a result, the model
        induces undirected relations among tokens as an intermediate
        step of learning representations.  We validate the performance
        of the LSTMN in language modeling, sentiment analysis, and
        natural language inference. In all cases, we train LSTMN
        models end-to-end with task-specific supervision signals,
        achieving performance comparable or better to state-of-the-art
        models and superior to vanilla LSTMs.
	
	\section{Related Work}
	\label{sec:related-work}
       
        Our machine reader is a recurrent neural network exhibiting
        two important properties: it is incremental, simulating human
        behavior, and performs shallow structure reasoning over input
        streams.  

        Recurrent neural network (RNNs) have been successfully applied
        to various sequence modeling and sequence-to-sequence
        transduction tasks. The latter have assumed several guises in
        the literature such as machine translation
        \cite{bahdanau2014neural}, sentence compression
        \cite{rush2015neural}, and reading comprehension
        \cite{hermann2015teaching}.  A key contributing factor to
        their success has been the ability to handle well-known
        problems with exploding or vanishing gradients
        \cite{bengio1994learning},  leading to models with
        gated activation functions
        \cite{hochreiter1997long,cho2014learning}, and more advanced
        architectures that enhance the information flow within the
        network
        \cite{koutnik2014clockwork,chung2015gated,yao2015depth}.
        
        A remaining practical bottleneck for RNNs is memory
        compression \cite{bahdanau2014neural}: since the inputs are
        recursively combined into a single memory representation which
        is typically too small in terms of parameters, it becomes
        difficult to accurately memorize sequences
        \cite{zaremba2014learning}. In the encoder-decoder
        architecture, this problem can be sidestepped with an
        attention mechanism which learns soft alignments
        \textit{between} the decoding states and the encoded memories
        \cite{bahdanau2014neural}. In our model, memory and attention
        are added \textit{within} a sequence encoder allowing the
        network to uncover lexical relations between tokens.


        The idea of introducing a structural bias to neural models is
        by no means new. For example, it is reflected in the work of
        \newcite{socher-EtAl:2013:EMNLP} who apply recursive neural
        networks for learning natural language representations.  In
        the context of recurrent neural networks, efforts to build
        modular, structured neural models date back to
        \newcite{das1992learning} who connect a recurrent neural
        network with an external memory stack for learning context
        free grammars. Recently, \newcite{weston2014memory} propose
        Memory Networks to explicitly segregate memory storage from
        the computation of neural networks in general. Their model is
        trained end-to-end with a memory addressing mechanism closely
        related to soft attention \cite{sukhbaatar2015end} and has
        been applied to machine translation \cite{meng2016deep}.
        \newcite{grefenstette2015learning} define a set of
        differentiable data structures (stacks, queues, and dequeues)
        as memories controlled by a recurrent neural
        network. \newcite{ke2016memory} combine the LSTM with an
        external memory block component which interacts with its
        hidden state.  \newcite{kumar2015ask} employ a structured
        neural network with episodic memory modules for natural
        language and also visual question answering
        \cite{xiong2016dynamic}.
        
        Similar to the above work, we leverage memory and attention in
        a recurrent neural network for inducing relations between
        tokens as a module in a larger network responsible for
        representation learning. As a property of soft attention, all
        intermediate relations we aim to capture are soft and
        differentiable. This is in contrast to shift-reduce type
        neural models \cite{dyer2015transition,bowman2016fast} where
        the intermediate decisions are hard and induction is more
        difficult.  Finally, note that our model captures undirected
        lexical relations and is thus distinct from work on dependency
        grammar induction \cite{klein-manning:2004:ACL} where the
        learned head-modifier relations are directed.
       
	
	\section{The Machine Reader}
	\label{sec:machine-reader}
	
        In this section we present our machine reader which is
        designed to process structured input while retaining the
        incrementality of a recurrent neural network.  The core of our
        model is a Long Short-Term Memory (LSTM) unit with an extended
        memory tape that explicitly simulates the human memory
        span. The model performs implicit relation analysis between
        tokens with an attention-based memory addressing mechanism at
        every time step. In the following, we first review the
        standard Long Short-Term Memory and then describe our model.
	
	\subsection{Long Short-Term Memory}
	\label{sec:long-short-term-1}
	A Long Short-Term Memory (LSTM) recurrent neural network processes a
	variable-length sequence $x=(x_1, x_2, \cdots, x_n)$ by incrementally
	adding new content into a single memory slot, with gates controlling
	the extent to which new content should be memorized, old content
	should be erased, and current content should be exposed. At time
	step~$t$, the memory~$c_t$ and the hidden state~$h_t$ are updated with
	the following equations:
	\begin{equation}
		\begin{bmatrix}
			i_t\\ f_t\\ o_t\\ \hat{c}_t
		\end{bmatrix} =
		\begin{bmatrix} \sigma\\ \sigma\\ \sigma\\ \tanh
		\end{bmatrix} W\cdot [h_{t-1}, \, x_t]
		\label{beginlstm}
	\end{equation}
	\begin{equation} c_t = f_t \odot c_{t-1} +
		i_t \odot \hat{c}_t
	\end{equation}
	\begin{equation} h_t = o_t \odot \tanh(c_t)
		\label{endlstm}
	\end{equation} 
	where $i$, $f$, and $o$ are gate activations. Compared to the
        standard RNN, the LSTM uses additive memory updates and it separates the memory~$c$ from the
        hidden state~$h$, which interacts with the environment when
        making predictions.

	\subsection{Long Short-Term Memory-Network}
	\label{sec:long-short-term}
	The first question that arises with LSTMs is the extent to
        which they are able to memorize sequences under recursive
        compression.  LSTMs can produce a list of state
        representations during composition, however, the next state is
        always computed from the current state. That is to say, given
        the current state $h_{t}$, the next state $h_{t+1}$ is
        conditionally independent of states $h_{1}\cdots h_{t-1}$ and
        tokens $x_{1} \cdots x_{t}$.  While the recursive state update
        is performed in a Markov manner, it is assumed that LSTMs
        maintain unbounded memory (i.e.,~the current state alone
        summarizes well the tokens it has seen so far).  This
        assumption may fail in practice, for example when the sequence is
        long or when the memory size is not large enough.  Another
        undesired property of LSTMs concerns modeling structured
        input. An LSTM aggregates information on a token-by-token
        basis in sequential order, but there is no explicit mechanism for
        reasoning over structure and modeling relations between
        tokens.


        Our model aims to address both limitations. Our solution is to
        modify the standard LSTM structure by replacing the memory
        cell with a memory network \cite{weston2014memory}.  The
        resulting Long Short-Term Memory-Network (LSTMN) stores the
        contextual representation of each input token with a unique
        memory slot and the size of the memory grows with time until
        an upper bound of the memory span is reached.  This design
        enables the LSTM to reason about relations between tokens with
        a neural attention layer and then perform non-Markov state
        updates.  Although it is feasible to apply both write and read
        operations to the memories with attention, we concentrate on
        the latter. We conceptualize the \emph{read} operation as
        attentively linking the current token to previous memories and
        selecting useful content when processing it. Although not the
        focus of this work, the significance of the \emph{write}
        operation can be analogously justified as a way of
        incrementally updating previous memories, e.g.,~to correct
        wrong interpretations when processing garden path sentences
        \cite{Ferreira:Henderson:1991}.

			\begin{figure}[t]
				\begin{center}
					\hspace*{-1.5ex}\includegraphics[width=0.5\textwidth]{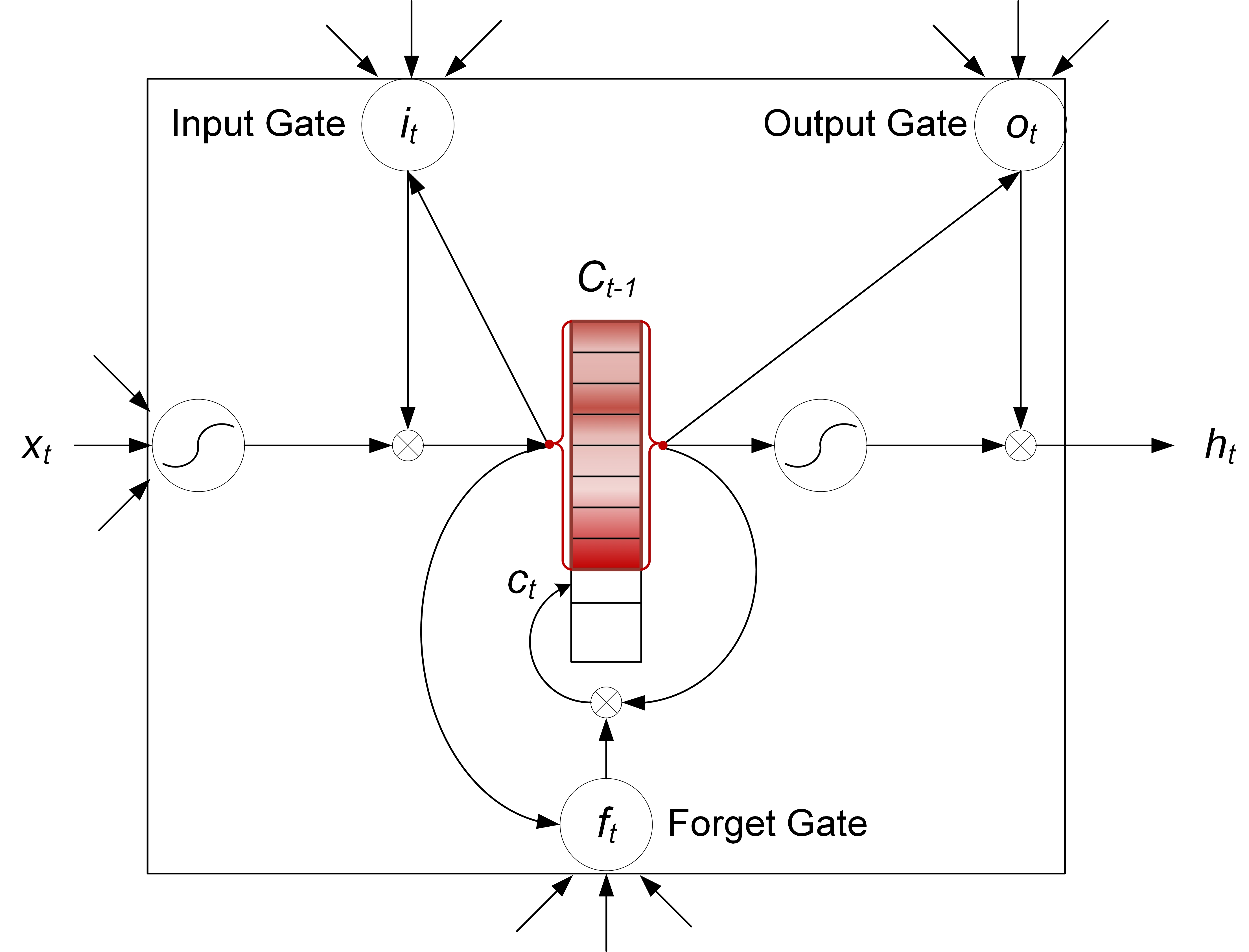}
				\end{center}
				\caption{\label{lstmn} Long Short-Term Memory-Network. Color indicates degree of memory activation.}
				\vspace{-2.5ex}
			\end{figure}

                        The architecture of the LSTMN is shown in
                        Figure~\ref{lstmn} and the formal definition
                        is provided as follows.  The model maintains
                        two sets of vectors stored in a hidden state
                        tape used to interact with the environment
                        (e.g.,~computing attention), and a memory tape
                        used to represent what is actually stored in
                        memory.\footnote{For comparison, LSTMs maintain a hidden
                          vector and a memory vector; memory networks
                          \cite{weston2014memory} have a set of key
                          vectors and a set of value vectors.}
                        Therefore, each token is associated with a
                        hidden vector and a memory vector.  Let~$x_t$
                        denote the current input; $C_{t-1} = (c_1,
                        \cdots, c_{t-1})$ denotes the current memory
                        tape, and $H_{t-1} = (h_1, \cdots, h_{t-1})$
                        the previous hidden tape.  At time step~$t$,
                        the model computes the relation between~$x_t$
                        and~$x_1\cdots x_{t-1}$ through~$h_1 \cdots
                        h_{t-1}$ with an attention layer:
                        \begin{equation}
                        a_i^t = v^\text{T} \tanh(W_h h_i + W_x x_t + W_{\tilde{h}} \tilde{h}_{t-1})
                        \label{intraatt}
                        \end{equation}
                        \begin{equation}
                        s_i^t = \text{softmax} (a_i^t)
                        \label{softmax}
                        \end{equation}
                        This yields a probability distribution over the hidden state vectors of previous tokens.
                        We can then compute an adaptive summary vector for the previous hidden tape and memory tape
                        denoted by~$\tilde{c}_t$ and~$\tilde{h}_t$, respectively:
                        \begin{equation}
                        \begin{bmatrix}
                        \tilde{h}_t\\ \tilde{c}_t
                        \end{bmatrix} = \sum\limits_{i=1}^{t-1} s_i^t \cdot
                        \begin{bmatrix}
                        h_i\\ c_i
                        \end{bmatrix} 
                        \end{equation}
                        and use them for computing the
                        values of~$c_t$ and~$h_t$ in the recurrent update as:
	\begin{equation}
		\begin{bmatrix}
			i_t\\ f_t\\ o_t\\ \hat{c}_t
		\end{bmatrix} =
		\begin{bmatrix} \sigma\\ \sigma\\ \sigma\\ \tanh
		\end{bmatrix} W \cdot [\tilde{h}_t, \, x_t]
		\label{gates}
	\end{equation}
	\begin{equation} c_t = f_t \odot \tilde{c}_t +
		i_t \odot \hat{c}_t
	\end{equation}
	\begin{equation} h_t = o_t \odot \tanh(c_t)
	\label{endintra}
	\end{equation} 
	where $v$, $W_h$, $W_x$ and $W_{\tilde{h}}$ are the new weight terms of the network. 
	

	A key idea behind the LSTMN is to use attention for inducing
        relations between tokens.  These relations are soft and
        differentiable, and components of a larger representation
        learning network. Although it is appealing to provide direct
        supervision for the attention layer, e.g.,~with evidence
        collected from a dependency treebank, we treat it as a
        submodule being optimized within the larger network in a
        downstream task.  It is also possible to have a more
        structured relational reasoning module by stacking multiple
        memory and hidden layers in an alternating fashion, resembling
        a stacked LSTM \cite{graves2013generating} or a multi-hop
        memory network \cite{sukhbaatar2015end}.  This can be achieved
        by feeding the output~${h}_t^k$ of the lower layer~$k$ as
        input to the upper layer~$(k+1)$.  The attention at the
        $(k+1)$th layer is computed as:
        \begin{equation}
        a_{i, k+1}^t = v^\text{T} \tanh(W_h h_{i} ^{k+1} + W_l {h}_t^k + W_{\tilde{h}} \tilde{h}_{t-1} ^ {k+1})
        \label{intraatt2}
        \end{equation}
        Skip-connections \cite{graves2013generating} can be applied to feed
        $x_t$ to upper layers as well.

	\begin{figure*}[t]
		\begin{center}
			\subfloat[Decoder with shallow attention fusion.] {
\includegraphics[width=0.44\linewidth]{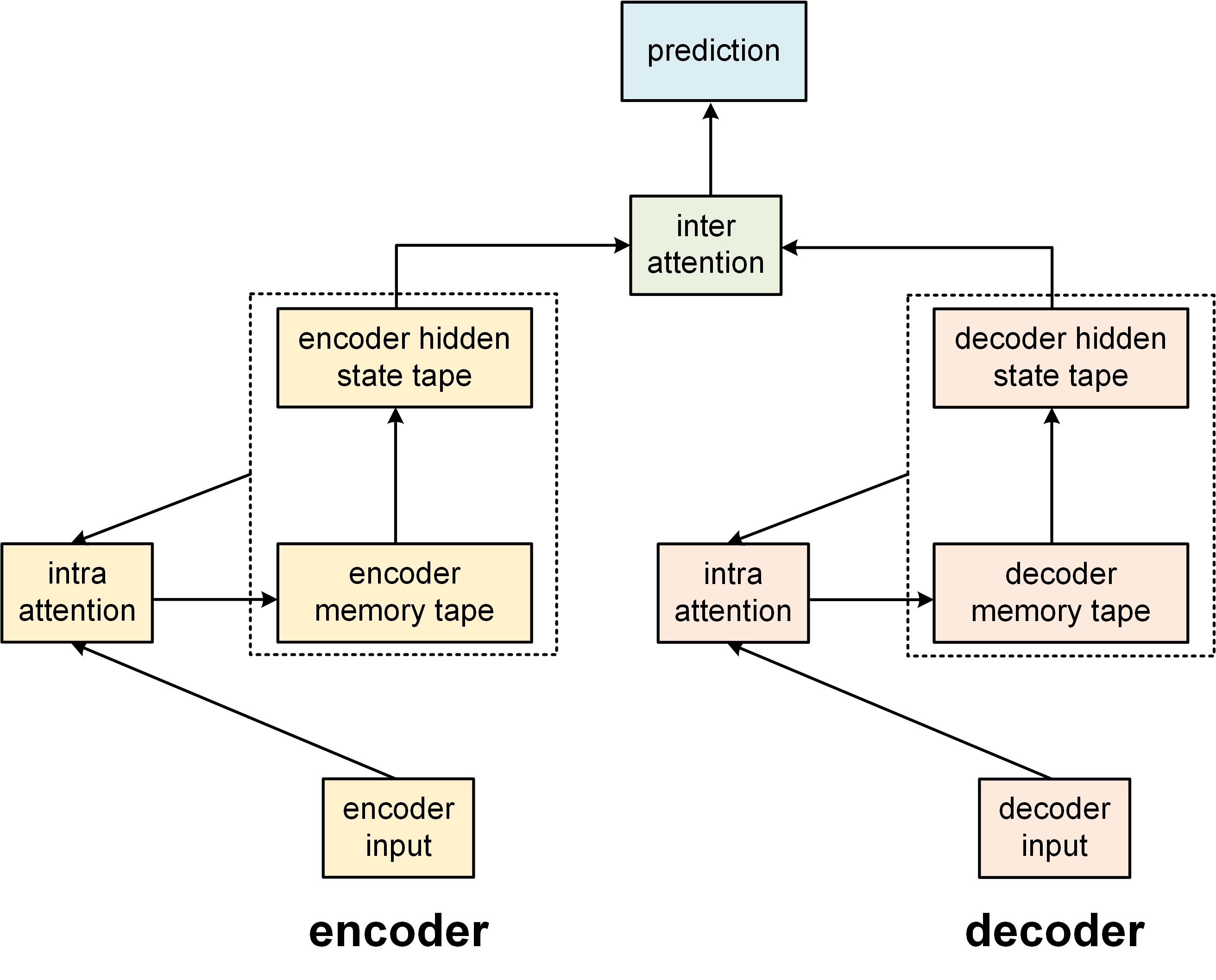} }
			\label{shallow} \quad
			\subfloat[Decoder with deep attention fusion.] {
			\includegraphics[width=0.5\linewidth]{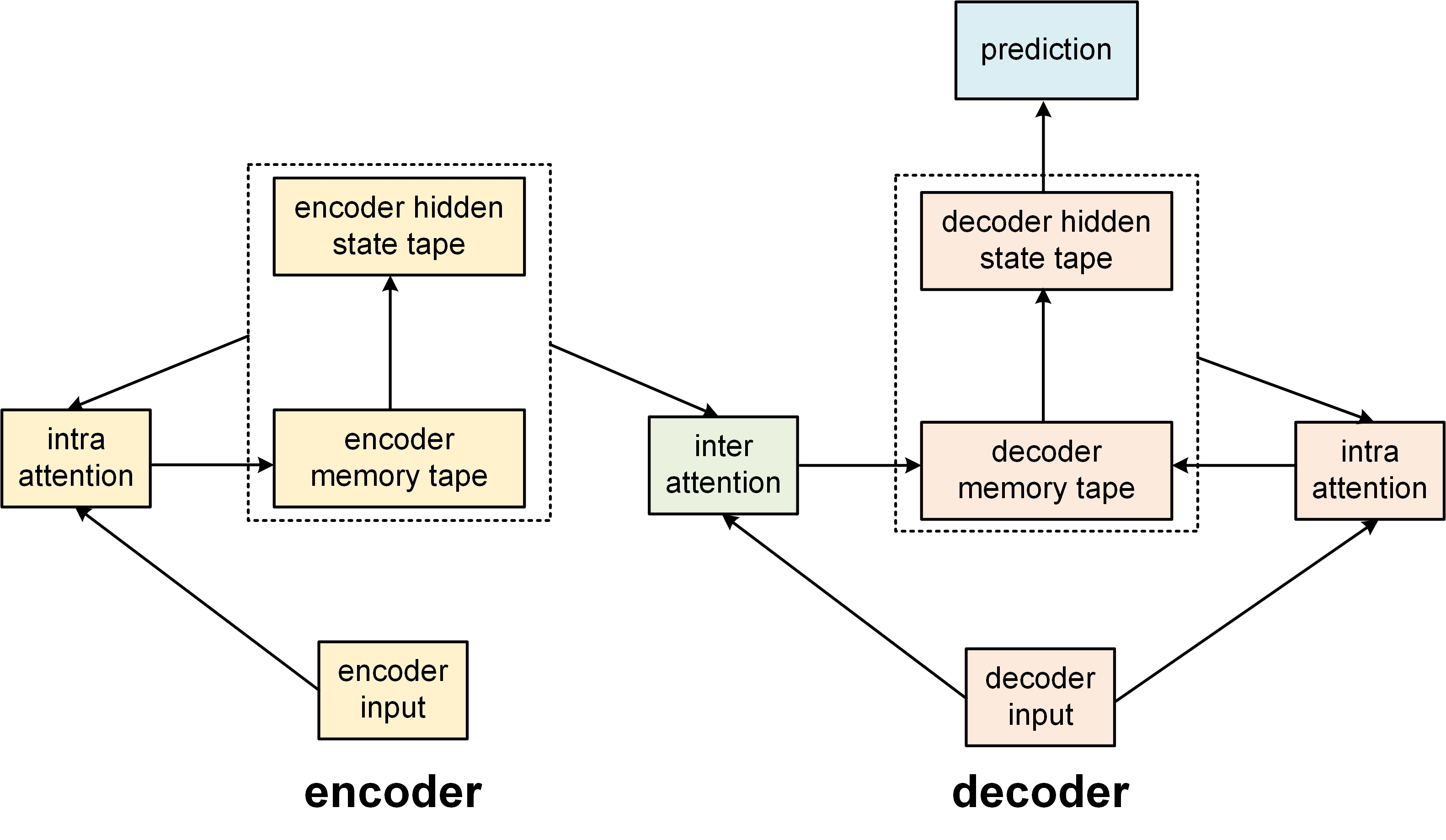}
			}
			\label{deep}
		\end{center}
		\caption{LSTMNs for sequence-to-sequence modeling. The
                  encoder uses intra-attention, while the decoder
                  incorporates both intra- and inter-attention. The
                  two figures present two ways to combine the intra-
                  and inter-attention in the decoder. }
		\label{att}
		\vspace{-2.5ex}
	\end{figure*}

	\section{Modeling Two Sequences with LSTMN}
	\label{sec:lstmns-dual-sequence}
	
	Natural language processing tasks such as machine translation
        and textual entailment are concerned with modeling two
        sequences rather than a single one. A standard tool for
        modeling two sequences with recurrent networks is the
        encoder-decoder architecture where the second sequence (also
        known as the \textit{target}) is being processed conditioned
        on the first one (also known as the \textit{source}). In this
        section we explain how to combine the LSTMN which applies
        attention for intra-relation reasoning, with the
        encoder-decoder network whose attention module learns the
        inter-alignment between two sequences. Figures~\ref{att}a and
        \ref{att}b illustrate two types of combination. We describe the
        models more formally below. 
	
	\paragraph{Shallow Attention Fusion}      
        Shallow fusion simply treats the LSTMN as a separate module
        that can be readily used in an encoder-decoder architecture,
        in lieu of a standard RNN or LSTM. As shown in
        Figure~\ref{att}a, both encoder and decoder are modeled as
        LSTMNs with intra-attention. Meanwhile, inter-attention is
        triggered when the decoder reads a target token, similar to
        the inter-attention introduced in
        \newcite{bahdanau2014neural}.
	
	\paragraph{Deep Attention Fusion} Deep fusion combines inter-
        and intra-attention (initiated by the decoder) when computing
        state updates.  We use different notation to represent the two
        sets of attention.  Following
        Section~\ref{sec:long-short-term}, $C$ and~$H$ denote the
        target memory tape and hidden tape, which store
        representations of the target symbols that have been processed
        so far.  The computation of intra-attention follows
        Equations~\eqref{intraatt}--\eqref{endintra}.  Additionally, we
        use $A=[\alpha_1, \cdots, \alpha_m]$ and $Y=[\gamma_1, \cdots,
        \gamma_m]$ to represent the source memory tape and hidden
        tape, with~$m$ being the length of the source sequence
        conditioned upon. We compute inter-attention between the input
        at time step~$t$ and tokens in the entire source sequence as
        follows:
	\begin{equation}
		b_j^t = u^\text{T} \tanh(W_{\gamma} {\gamma}_j + W_x x_t + W_{\tilde{\gamma}} \tilde{\gamma}_{t-1})
	\end{equation}
	\begin{equation}
		p_j^t = \text{softmax} (b_j^t)
	\end{equation}
	After that we compute the adaptive representation of the source memory
        tape $\tilde{\alpha}_t$ and hidden tape $\tilde{\gamma}_t$ as:
	\begin{equation}
		\begin{bmatrix}
			\tilde{\gamma}_t \\ \tilde{\alpha}_t
		\end{bmatrix} = \sum\limits_{j=1}^{m} p_j^t \cdot
		\begin{bmatrix}
			\gamma_j \\ \alpha_j
		\end{bmatrix} 
	\end{equation}
	We can then transfer the adaptive source
        representation~$\tilde{\alpha}_t$ to the target memory with
        another gating operation $r_t$, analogous to the gates in
        Equation~\eqref{gates}.
		\begin{equation}
			r_t = \sigma (W_r \cdot [\tilde{\gamma}_t, x_t])
		\end{equation}
                The new target memory includes inter-alignment $r_t
                \odot \tilde{\alpha}_t$, intra-relation $f_t \odot
                \tilde{c}_{t}$, and the new input information $i_t
                \odot \hat{c}_t $:
	\begin{equation} c_t =  r_t \odot \tilde{\alpha}_t + f_t \odot \tilde{c}_{t} +
		i_t \odot \hat{c}_t 
	\end{equation}
	\begin{equation} h_t = o_t \odot \tanh(c_t)
	\end{equation} 
	As shown in the equations above and Figure~\ref{att}b, the
        major change of deep fusion lies in the recurrent storage of
        the inter-alignment vector in the target memory network, as a
        way to help the target network review source information.
	
	\section{Experiments}
	\label{sec:experiments}
	
	In this section we present our experiments for evaluating the
        performance of the LSTMN machine reader. We start with
        language modeling as it is a natural testbed for our model. We
        then assess the model's ability to extract meaning
        representations for generic sentence classification tasks such
        as sentiment analysis. Finally, we examine whether the LSTMN
        can recognize the semantic relationship between two sentences
        by applying it to a natural language inference task. Our code
        is available at
        \url{https://github.com/cheng6076/SNLI-attention}.
	
	\subsection{Language Modeling}
	\label{sec:language-modeling}
	

	\begin{table}[t]
		\centering
	    \small
		\begin{tabular}{|l|c|c|}
			\hline
			Models & Layers & Perplexity \\
			\hline\hline
			KN5 & --- & 141 \\\hline
			RNN & 1 & 129 \\
			LSTM & 1 & 115 \\
			LSTMN & 1 &  \textbf{108} \\
			\hline
			sLSTM & 3 &  115 \\ 
			gLSTM & 3 & 107 \\
			dLSTM & 3 & 109 \\
			LSTMN & 3 &  \textbf{102} \\
			\hline
		\end{tabular}
		\caption{Language model perplexity on 
                  the Penn Treebank. The size of memory is 300 for all
                  models.} 
		\label{lmr}
		\vspace{-2.5ex}
	\end{table}

	Our language modeling experiments were conducted on the
        English Penn Treebank dataset. Following common practice
        \cite{mikolov2010recurrent}, we trained on sections 0--20 (1M
        words), used sections 21--22 for validation (80K words), and
        sections 23--24 (90K words for testing). The dataset contains
        approximately 1~million tokens and a vocabulary size
        of~10K. The average sentence length is~21.
	We use perplexity as our evaluation metric: \mbox{$PPL =
          \exp(NLL/T)$}, where $NLL$ denotes the negative log
        likelihood of the entire test set and~$T$ the corresponding
        number of tokens. We used stochastic gradient descent for
        optimization with an initial learning rate of~0.65, which
        decays by a factor of~0.85 per epoch if no significant
        improvement has been observed on the validation set. We
        renormalize the gradient if its norm is greater than~5. The
        \mbox{mini-batch} size was set to~40. The dimensions of the
        word embeddings were set to~150 for all models.

	In this suite of experiments we compared the LSTMN against a
        variety of baselines. The first one is a Kneser-Ney 5-gram
        language model (KN5) which generally serves as a non-neural
        baseline for the language modeling task. We also present
        perplexity results for the standard RNN and LSTM models. We
        also implemented more sophisticated LSTM architectures, such
        as a stacked LSTM (sLSTM), a gated-feedback LSTM (gLSTM;
        \newcite{chung2015gated}) and a depth-gated LSTM (dLSTM;
        \newcite{yao2015depth}). The gated-feedback LSTM has feedback
        gates connecting the hidden states across multiple time steps
        as an adaptive control of the information flow. The
        depth-gated LSTM uses a depth gate to connect memory cells of
        vertically adjacent layers. In general, both gLSTM and
        \mbox{dLSTM} are able to capture long-term dependencies to
        some degree, but they do not explicitly keep past memories.
        We set the number of layers to~3 in this experiment, mainly to
        agree with the language modeling experiments of
        \newcite{chung2015gated}. Also note that that there are no
        single-layer variants for gLSTM and dLSTM; they have to
        be implemented as multi-layer systems. The hidden unit size of
        the LSTMN and all comparison models (except KN5) was set
        to~300.

\begin{figure}[t]
		\centering
		\begin{center}
			\includegraphics[width=0.4\textwidth]{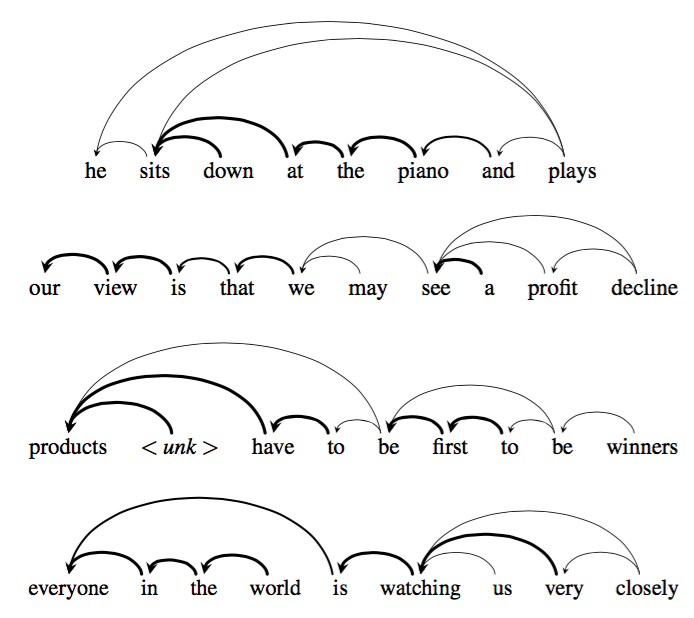}
		\end{center}
		\caption{\label{vis4lm} Examples of intra-attention
			(language modeling). Bold lines indicate higher attention
			scores. Arrows denote which word is being focused when
			attention is computed, but not the direction of the relation.}
		\vspace{-2.5ex}
	\end{figure}

	The results of the language modeling task are shown in
        Table~\ref{lmr}. Perplexity results for KN5 and RNN are taken
        from \newcite{mikolov2015learning}. As can be seen, the
        single-layer LSTMN outperforms these two baselines and the
        LSTM by a significant margin. Amongst all deep architectures,
        the three-layer LSTMN also performs best. We can study the
        memory activation mechanism of the machine reader by
        visualizing the attention scores. Figure~\ref{vis4lm} shows
        four sentences sampled from the Penn Treebank validation set.
        Although we explicitly encourage the reader to attend to any
        memory slot, much attention focuses on recent memories. This
        agrees with the linguistic intuition that long-term
        dependencies are relatively rare. As illustrated in
        Figure~\ref{vis4lm} the model captures some valid lexical
        relations (e.g., the dependency between \textsl{sits} and
        \textsl{at}, \textsl{sits} and \textsl{plays},
        \textsl{everyone} and \textsl{is}, \textsl{is} and
        \textsl{watching}).  Note that arcs here are undirected and are different from
        the directed arcs denoting head-modifier
        relations in dependency graphs.
	
	\subsection{Sentiment Analysis}
	\label{sec:sentiment-analysis}
	
	Our second task concerns the prediction of sentiment labels of
        sentences. We used the Stanford Sentiment Treebank
        \cite{socher-EtAl:2013:EMNLP}, which contains fine-grained
        sentiment labels (very positive, positive, neutral, negative,
        very negative) for 11,855 sentences.  Following previous work
        on this dataset, we used 8,544 sentences for training, 1,101
        for validation, and 2,210 for testing. The average sentence
        length is 19.1.  In addition, we also performed a binary
        classification task (positive, negative) after removing the
        neutral label. This resulted in 6,920 sentences for training,
        872 for validation and 1,821 for
        testing. Table~\ref{sentiment} reports results on both
        fine-grained and binary classification tasks.

	We experimented with 1- and 2-layer LSTMNs.  For the latter
        model, we predict the sentiment label of the sentence based on
        the averaged hidden vector passed to a 2-layer neural network
        classifier with ReLU as the activation function.  The memory
        size for both LSTMN models was set to 168 to be compatible
        with previous LSTM models \cite{tai2015improved} applied to
        the same task.  We used pre-trained \texttt{300-D Glove 840B}
        vectors \cite{pennington2014glove} to initialize the word
        embeddings. The gradient for words with \texttt{Glove}
        embeddings, was scaled by~0.35 in the first epoch after which
        all word embeddings were updated normally.

        We used Adam \cite{kingma2014adam} for optimization with the
        two momentum parameters set to~0.9 and~0.999 respectively. The
        initial learning rate was set to 2E-3. The regularization
        constant was 1E-4 and the mini-batch size was~5. A dropout
        rate of~0.5 was applied to the neural network classifier.
	
\begin{table}[t]
		\centering
		\label{my-label}
		\small
		\begin{tabular}{|@{~}l@{~}|@{~}c@{~}|@{~}c@{~}|}
			\hline
			Models & Fine-grained& Binary \\
			\hline
			RAE {\scriptsize \cite{socher2011dynamic}} & 43.2 & 82.4 \\
			RNTN {\scriptsize \cite{socher2013recursive}} & 45.7 & 85.4 \\
			DRNN {\scriptsize \cite{irsoy2014deep}} & 49.8 &  86.6 \\ 
			\hline
			DCNN {\scriptsize \cite{blunsom2014convolutional}} & 48.5 & 86.8 \\
			CNN-MC {\scriptsize \cite{kim2014convolutional}} & 48.0 & 88.1 \\
			T-CNN {\scriptsize \cite{lei-barzilay-jaakkola:2015:EMNLP}} & \textbf{51.2} &
			\textbf{88.6} \\
			PV {\scriptsize \cite{le2014distributed}} & 48.7 &  87.8 \\
			\hline
			CT-LSTM {\scriptsize \cite{tai2015improved}} & 51.0 & 88.0 \\
			LSTM {\scriptsize \cite{tai2015improved}} & 46.4 & 84.9 \\
			2-layer LSTM {\scriptsize \cite{tai2015improved}} & 46.0 & 86.3 \\
			\textbf{LSTMN} & \textbf{47.6} & \textbf{86.3}\\
			\textbf{2-layer LSTMN} & \textbf{47.9} & \textbf{87.0}\\
			\hline
		\end{tabular}
		\caption{Model accuracy (\%) on the Sentiment
                  Treebank (test set). The memory size of LSTMN models
                  is set to 168 to be compatible with previously
                  published LSTM variants (Tai et al., 2015).}
		\label{sentiment}
		\vspace{-2.5ex}
	\end{table}

	We compared our model with a wide range of top-performing systems.
	Most of these models (including ours) are LSTM variants (third block
	in Table~\ref{sentiment}), recursive neural networks (first block), or
	convolutional neural networks (CNNs; second block). Recursive models
	assume the input sentences are represented as parse trees and can take
	advantage of annotations at the phrase level.  LSTM-type models and
	CNNs are trained on sequential input, with the exception of
	\mbox{CT-LSTM} \cite{tai2015improved} which operates over
	tree-structured network topologies such as constituent trees.  For
	comparison, we also report the performance of the paragraph vector
	model (PV; \newcite{le2014distributed}; see Table~\ref{sentiment},
	second block) which neither operates on trees nor sequences but learns
	distributed document representations parameterized directly.

\begin{figure}[t]
		\begin{center}
			\includegraphics[width=0.4\textwidth]{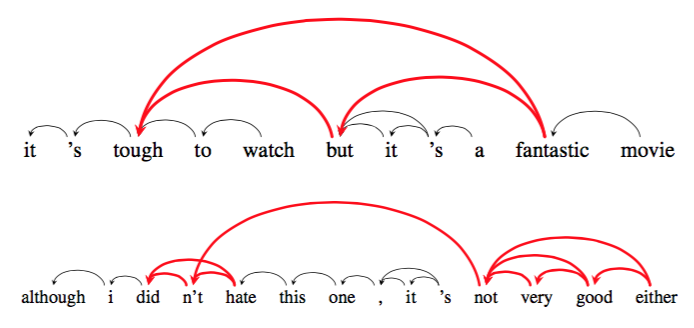}
		\end{center}
		\caption{\label{sa} Examples of intra-attention
			(sentiment analysis). Bold lines (red) indicate
			attention between sentiment important words.}
	\end{figure}

	The results in Table~\ref{sentiment} show that both 1- and
        2-layer LSTMNs outperform the LSTM baselines while achieving
        numbers comparable to state of the art.  The number of layers
        for our models was set to be comparable to previously
        published results. On the fine-grained and binary
        classification tasks our 2-layer LSTMN performs close to the
        best system T-CNN \cite{lei-barzilay-jaakkola:2015:EMNLP}.
        Figure~\ref{sa} shows examples of intra-attention for
        sentiment words. Interestingly, the network learns to
        associate sentiment important words such as \textsl{though}
        and \textsl{fantastic} or \textsl{not} and \textsl{good}.

	\subsection{Natural Language Inference}
	\label{sec:natur-lang-infer}
	
	The ability to reason about the semantic relationship between
        two sentences is an integral part of text understanding. We
        therefore evaluate our model on recognizing textual
        entailment, i.e.,~whether two premise-hypothesis pairs are
        entailing, contradictory, or neutral.  For this task we used
        the Stanford Natural Language Inference (SNLI) dataset
        \cite{bowman2015large}, which contains premise-hypothesis
        pairs and target labels indicating their relation. After
        removing sentences with unknown labels, we end up with~549,367
        pairs for training, 9,842 for development and 9,824 for
        testing. The vocabulary size is~36,809 and the average
        sentence length is~22. We performed lower-casing and tokenization for the entire dataset.
	
	Recent approaches use two sequential LSTMs to encode the
        premise and the hypothesis respectively, and apply neural
        attention to reason about their logical relationship
        \cite{rocktaschel2015reasoning,wang2015learning}. Furthermore,
        \newcite{rocktaschel2015reasoning} show that a non-standard
        encoder-decoder architecture which processes the hypothesis
        conditioned on the premise results significantly boosts
        performance.  We use a similar approach to tackle this task
        with LSTMNs.  Specifically, we use two LSTMNs to read the
        premise and hypothesis, and then match them by comparing their
        hidden state tapes. We perform average pooling for the hidden
        state tape of each LSTMN, and concatenate the two averages to
        form the input to a 2-layer neural network classifier with
        ReLU as the activation function.

	We used pre-trained \texttt{300-D Glove 840B} vectors
        \cite{pennington2014glove} to initialize the word
        embeddings. Out-of-vocabulary (OOV) words were initialized
        randomly with Gaussian samples ($\mu$=0, $\sigma$=1). We only
        updated OOV vectors in the first epoch, after which all word
        embeddings were updated normally.  The dropout rate was
        selected from [0.1, 0.2, 0.3, 0.4].  We used Adam
        \cite{kingma2014adam} for optimization with the two momentum
        parameters set to 0.9 and 0.999 respectively, and the initial
        learning rate set to~1E-3. The mini-batch size was set to~16
        or~32.  For a fair comparison against previous work, we report
        results with different hidden/memory dimensions (i.e.,~100,
        300, and 450).
	

	We compared variants of our model against different types of
        LSTMs (see the second block in
        Table~\ref{entailment}). Specifically, these include a model
        which encodes the premise and hypothesis independently with
        two LSTMs \cite{bowman2015large}, a shared LSTM
        \cite{rocktaschel2015reasoning}, a word-by-word attention
        model \cite{rocktaschel2015reasoning}, and a matching LSTM
        (mLSTM; \newcite{wang2015learning}). This model sequentially
        processes the hypothesis, and at each position tries to match
        the current word with an attention-weighted representation of
        the premise (rather than basing its predictions on whole
        sentence embeddings).  We also compared our models with a
        bag-of-words baseline which averages the pre-trained
        embeddings for the words in each sentence and
        concatenates them to create features for a logistic regression
        classifier (first block in Table~\ref{entailment}).  

        LSTMNs achieve better performance compared to LSTMs (with and
        without attention; 2nd block in Table~\ref{entailment}).  We
        also observe that fusion is generally beneficial, and that
        deep fusion slightly improves over shallow fusion. One
        explanation is that with deep fusion the inter-attention
        vectors are recurrently memorized by the decoder with a gating
        operation, which also improves the information flow of the
        network. With standard training, our deep fusion yields the
        state-of-the-art performance in this task. Although
        encouraging, this result should be interpreted with caution
        since our model has substantially more parameters compared to
        related systems. We could compare different models using the
        same number of total parameters. However, this would
        inevitably introduce other biases, e.g.,~the number of
        hyper-parameters would become different.


	\begin{table}[t]
		\centering
		\begin{tabular}{|@{~}l c  c  c@{~}|}
			\hline
			Models & $h$ & $|\theta|_{\text{M}}$ & Test \\\hline
			\hline
			 BOW concatenation & --- &   ---  & 59.8 \\
			\hline
			LSTM {\scriptsize  \cite{bowman2015large}}  & 100 &  221k  & 77.6 \\
			LSTM-att {\scriptsize \cite{rocktaschel2015reasoning}} & 100 & 252k &  83.5 \\
			 mLSTM {\scriptsize \cite{wang2015learning}} & 300 & 1.9M &  {86.1} \\     
			\hline
			LSTMN & 100 & 260k  & 81.5 \\
			LSTMN shallow fusion & 100 & 280k  &  84.3\\
			LSTMN deep fusion & 100 & 330k  &  84.5 \\
			LSTMN shallow fusion & 300 & 1.4M  &  85.2 \\
			LSTMN deep fusion & 300 & 1.7M &  85.7 \\
			LSTMN shallow fusion & 450 & 2.8M   &  86.0\\
			LSTMN deep fusion & 450 & 3.4M & \textbf{86.3} \\
			\hline
		\end{tabular}
		\caption{Parameter counts $|\theta|_{\text{M}}$, size
                  of hidden unit~$h$, and  model accuracy (\%) 
                  on the natural language inference task.} 
		\label{entailment}
		\vspace{-2.5ex}
	\end{table}

	\section{Conclusions}
	\label{sec:conclusions}
	
	In this paper we proposed a machine reading simulator to
        address the limitations of recurrent neural networks when
        processing inherently structured input.  Our model is based on
        a Long Short-Term Memory architecture embedded with a memory
        network, explicitly storing contextual representations of
        input tokens without recursively compressing them.  More
        importantly, an intra-attention mechanism is employed for
        memory addressing, as a way to induce undirected relations
        among tokens. The attention layer is not optimized with a
        direct supervision signal but with the entire network in
        downstream tasks.  Experimental results across three tasks
        show that our model yields performance comparable or superior
        to state of the art.
	
	Although our experiments focused on LSTMs, the idea of
        building more structure aware neural models is general and can
        be applied to other types of networks. When direct supervision
        is provided, similar architectures can be adapted to tasks
        such as dependency parsing and relation extraction. In the
        future, we hope to develop more linguistically plausible
        neural architectures able to reason over nested structures and
        neural models that learn to discover compositionality with
        weak or indirect supervision.
	
	\section*{Acknowledgments}
        We thank members of the ILCC at the School of Informatics and
        the anonymous reviewers for helpful comments. The support of
        the European Research Council under award number 681760
        ``Translating Multiple Modalities into Text'' is gratefully
        acknowledged.

	\bibliographystyle{emnlp2016}
	\bibliography{emnlp2016,mirella}

\begin{thebibliography}{}

\bibitem[\protect\citename{Andreas \bgroup et al.\egroup
  }2016]{andreas2016learning}
Jacob Andreas, Marcus Rohrbach, Trevor Darrell, and Dan Klein.
\newblock 2016.
\newblock Learning to compose neural networks for question answering.
\newblock In {\em Proceedings of the 2016 NAACL: HLT}, pages 1545--1554, San
  Diego, California.

\bibitem[\protect\citename{Bahdanau \bgroup et al.\egroup
  }2014]{bahdanau2014neural}
Dzmitry Bahdanau, Kyunghyun Cho, and Yoshua Bengio.
\newblock 2014.
\newblock Neural machine translation by jointly learning to align and
  translate.
\newblock In {\em Proceedings of the 2014 ICLR}, Banff, Alberta.

\bibitem[\protect\citename{Bengio \bgroup et al.\egroup
  }1994]{bengio1994learning}
Yoshua Bengio, Patrice Simard, and Paolo Frasconi.
\newblock 1994.
\newblock Learning long-term dependencies with gradient descent is difficult.
\newblock {\em Neural Networks, IEEE Transactions on}, 5(2):157--166.

\bibitem[\protect\citename{Blunsom \bgroup et al.\egroup
  }2014]{blunsom2014convolutional}
Phil Blunsom, Edward Grefenstette, and Nal Kalchbrenner.
\newblock 2014.
\newblock A convolutional neural network for modelling sentences.
\newblock In {\em Proceedings of the 52nd ACL}, pages 655--665, Baltimore,
  Maryland.

\bibitem[\protect\citename{Bowman \bgroup et al.\egroup }2015]{bowman2015large}
Samuel~R Bowman, Gabor Angeli, Christopher Potts, and Christopher~D Manning.
\newblock 2015.
\newblock A large annotated corpus for learning natural language inference.
\newblock In {\em Proceedings of the 2015 EMNLP}, pages 22--32, Lisbon,
  Portugal.

\bibitem[\protect\citename{Bowman \bgroup et al.\egroup }2016]{bowman2016fast}
Samuel~R Bowman, Jon Gauthier, Abhinav Rastogi, Raghav Gupta, Christopher~D
  Manning, and Christopher Potts.
\newblock 2016.
\newblock A fast unified model for parsing and sentence understanding.
\newblock In {\em Proceedings of the 54th ACL}, pages 1466--1477, Berlin,
  Germany.

\bibitem[\protect\citename{Cho \bgroup et al.\egroup }2014]{cho2014learning}
Kyunghyun Cho, Bart Van~Merri{\"e}nboer, Caglar Gulcehre, Dzmitry Bahdanau,
  Fethi Bougares, Holger Schwenk, and Yoshua Bengio.
\newblock 2014.
\newblock Learning phrase representations using {RNN} encoder-decoder for
  statistical machine translation.
\newblock In {\em Proceedings of the 2014 EMNLP}, pages 1724--1734, Doha,
  Qatar.

\bibitem[\protect\citename{Chung \bgroup et al.\egroup }2015]{chung2015gated}
Junyoung Chung, Caglar Gulcehre, Kyunghyun Cho, and Yoshua Bengio.
\newblock 2015.
\newblock Gated feedback recurrent neural networks.
\newblock In {\em Proceedings of the 32nd ICML}, pages 2067--2075, Lille,
  France.

\bibitem[\protect\citename{Clark \bgroup et al.\egroup }2013]{Clark:ea:2013}
Peter Clark, Phil Harrison, and Niranjan Balasubramanian.
\newblock 2013.
\newblock A study of the knowledge base requirements for passing an elementary
  science test.
\newblock In {\em Proceedings of the 3rd Workshop on Automated KB
  Construction}, San Francisco, California.

\bibitem[\protect\citename{Dagan \bgroup et al.\egroup }2005]{Dagan:ea:2005}
Ido Dagan, Oren Glickman, and Bernardo Magnini.
\newblock 2005.
\newblock The {PASCAL} recognising textual entailment challenge.
\newblock In {\em Proceedings of the {PASCAL} Challenges Workshop on
  Recognising Textual Entailment}.

\bibitem[\protect\citename{Das \bgroup et al.\egroup }1992]{das1992learning}
Sreerupa Das, C.~Lee Giles, and Guo zheng Sun.
\newblock 1992.
\newblock Learning context-free grammars: Capabilities and limitations of a
  recurrent neural network with an external stack memory.
\newblock In {\em Proceedings of the 14th Annual Conference of the Cognitive
  Science Society}, pages 791--795. Morgan Kaufmann Publishers.

\bibitem[\protect\citename{Dyer \bgroup et al.\egroup
  }2015]{dyer2015transition}
Chris Dyer, Miguel Ballesteros, Wang Ling, Austin Matthews, and Noah~A Smith.
\newblock 2015.
\newblock Transition-based dependency parsing with stack long short-term
  memory.
\newblock In {\em Proceedings of the 53rd ACL}, pages 334--343, Beijing, China.

\bibitem[\protect\citename{Etzioni \bgroup et al.\egroup
  }2011]{Etzioni:etal:2011}
Oren Etzioni, Anthony Fader, Janara Christensen, Stephen Soderland, and Mausam.
\newblock 2011.
\newblock Open information extraction: The second generation.
\newblock In {\em Proceedings of the 22nd IJCAI}, pages 3--10, Barcelona,
  Spain.

\bibitem[\protect\citename{Fader \bgroup et al.\egroup
  }2011]{fader-soderland-etzioni:2011:EMNLP}
Anthony Fader, Stephen Soderland, and Oren Etzioni.
\newblock 2011.
\newblock Identifying relations for open information extraction.
\newblock In {\em Proceedings of the 2011 EMNLP}, pages 1535--1545, Edinburgh,
  Scotland, UK.

\bibitem[\protect\citename{Ferreira and
  Henderson}1991]{Ferreira:Henderson:1991}
Fernanda Ferreira and John~M. Henderson.
\newblock 1991.
\newblock Recovery from misanalyses of garden-path sentences.
\newblock {\em Journal of Memory and Language}, 30:725--745.

\bibitem[\protect\citename{Frank and Bod}2011]{Frank:Bod:2011}
Stefan~L. Frank and Rens Bod.
\newblock 2011.
\newblock Insensitivity of the human sentence-processing system to hierarchical
  structure.
\newblock {\em Pyschological Science}, 22(6):829--834.

\bibitem[\protect\citename{Graves}2013]{graves2013generating}
Alex Graves.
\newblock 2013.
\newblock Generating sequences with recurrent neural networks.
\newblock {\em arXiv preprint arXiv:1308.0850}.

\bibitem[\protect\citename{Grefenstette \bgroup et al.\egroup
  }2015]{grefenstette2015learning}
Edward Grefenstette, Karl~Moritz Hermann, Mustafa Suleyman, and Phil Blunsom.
\newblock 2015.
\newblock Learning to transduce with unbounded memory.
\newblock In {\em Advances in Neural Information Processing Systems}, pages
  1819--1827.

\bibitem[\protect\citename{Hermann \bgroup et al.\egroup
  }2015]{hermann2015teaching}
Karl~Moritz Hermann, Tomas Kocisky, Edward Grefenstette, Lasse Espeholt, Will
  Kay, Mustafa Suleyman, and Phil Blunsom.
\newblock 2015.
\newblock Teaching machines to read and comprehend.
\newblock In {\em Advances in Neural Information Processing Systems}, pages
  1684--1692.

\bibitem[\protect\citename{Hochreiter and Schmidhuber}1997]{hochreiter1997long}
Sepp Hochreiter and J{\"u}rgen Schmidhuber.
\newblock 1997.
\newblock Long short-term memory.
\newblock {\em Neural computation}, 9(8):1735--1780.

\bibitem[\protect\citename{Hochreiter}1991]{hochreiter1991untersuchungen}
Sepp Hochreiter.
\newblock 1991.
\newblock Untersuchungen zu dynamischen neuronalen netzen.
\newblock {\em Diploma, Technische Universit{\"a}t M{\"u}nchen}.

\bibitem[\protect\citename{Irsoy and Cardie}2014]{irsoy2014deep}
Ozan Irsoy and Claire Cardie.
\newblock 2014.
\newblock Deep recursive neural networks for compositionality in language.
\newblock In {\em Advances in Neural Information Processing Systems}, pages
  2096--2104.

\bibitem[\protect\citename{Kim}2014]{kim2014convolutional}
Yoon Kim.
\newblock 2014.
\newblock Convolutional neural networks for sentence classification.
\newblock In {\em Proceedings of the 2014 EMNLP}, pages 1746--1751, Doha,
  Qatar.

\bibitem[\protect\citename{Kingma and Ba}2015]{kingma2014adam}
Diederik Kingma and Jimmy Ba.
\newblock 2015.
\newblock Adam: A method for stochastic optimization.
\newblock In {\em Proceedings of the 2015 ICLR}, San Diego, California.

\bibitem[\protect\citename{Klein and Manning}2004]{klein-manning:2004:ACL}
Dan Klein and Christopher Manning.
\newblock 2004.
\newblock Corpus-based induction of syntactic structure: Models of dependency
  and constituency.
\newblock In {\em Proceedings of the 42nd ACL}, pages 478--485, Barcelona,
  Spain.

\bibitem[\protect\citename{Konieczny}2000]{Konieczny:2000}
Lars Konieczny.
\newblock 2000.
\newblock Locality and parsing complexity.
\newblock {\em Journal of Psycholinguistics}, 29(6):627--645.

\bibitem[\protect\citename{Koutn\'{i}k \bgroup et al.\egroup
  }2014]{koutnik2014clockwork}
Jan Koutn\'{i}k, Klaus Greff, Faustino Gomez, and J\"{u}rgen Schmidhuber.
\newblock 2014.
\newblock A clockwork {RNN}.
\newblock In {\em Proceedings of the 31st ICML}, pages 1863--1871, Beijing,
  China.

\bibitem[\protect\citename{Kumar \bgroup et al.\egroup }2016]{kumar2015ask}
Ankit Kumar, Ozan Irsoy, Jonathan Su, James Bradbury, Robert English, Brian
  Pierce, Peter Ondruska, Ishaan Gulrajani, and Richard Socher.
\newblock 2016.
\newblock Ask me anything: Dynamic memory networks for natural language
  processing.
\newblock In {\em Proceedings of the 33rd ICML}, New York, NY.

\bibitem[\protect\citename{Le and Mikolov}2014]{le2014distributed}
Quoc~V Le and Tomas Mikolov.
\newblock 2014.
\newblock Distributed representations of sentences and documents.
\newblock In {\em Proceedings of the 31st ICML}, pages 1188--1196, Beijing,
  China.

\bibitem[\protect\citename{Lei \bgroup et al.\egroup
  }2015]{lei-barzilay-jaakkola:2015:EMNLP}
Tao Lei, Regina Barzilay, and Tommi Jaakkola.
\newblock 2015.
\newblock Molding cnns for text: non-linear, non-consecutive convolutions.
\newblock In {\em Proceedings of the 2015 EMNLP}, pages 1565--1575, Lisbon,
  Portugal.

\bibitem[\protect\citename{Meng \bgroup et al.\egroup }2015]{meng2016deep}
Fandong Meng, Zhengdong Lu, Zhaopeng Tu, Hang Li, and Qun Liu.
\newblock 2015.
\newblock A deep memory-based architecture for sequence-to-sequence learning.
\newblock In {\em Proceedings of ICLR-Workshop 2016}, San Juan, Puerto Rico.

\bibitem[\protect\citename{Mikolov \bgroup et al.\egroup
  }2010]{mikolov2010recurrent}
Tomas Mikolov, Martin Karafi{\'a}t, Lukas Burget, Jan Cernock{\`y}, and Sanjeev
  Khudanpur.
\newblock 2010.
\newblock Recurrent neural network based language model.
\newblock In {\em Proceedings of 11th Interspeech}, pages 1045--1048, Makuhari,
  Japan.

\bibitem[\protect\citename{Mikolov \bgroup et al.\egroup
  }2015]{mikolov2015learning}
Tomas Mikolov, Armand Joulin, Sumit Chopra, Michael Mathieu, and Marc'Aurelio
  Ranzato.
\newblock 2015.
\newblock Learning longer memory in recurrent neural networks.
\newblock In {\em Proceedings of ICLR Workshop}, San Diego, California.

\bibitem[\protect\citename{Pascanu \bgroup et al.\egroup
  }2013]{pascanu2012difficulty}
Razvan Pascanu, Tomas Mikolov, and Yoshua Bengio.
\newblock 2013.
\newblock On the difficulty of training recurrent neural networks.
\newblock In {\em Proceedings of the 30th ICML}, pages 1310--1318, Atlanta,
  Georgia.

\bibitem[\protect\citename{Pennington \bgroup et al.\egroup
  }2014]{pennington2014glove}
Jeffrey Pennington, Richard Socher, and Christopher~D. Manning.
\newblock 2014.
\newblock Glove: Global vectors for word representation.
\newblock In {\em Proceedings of the 2014 EMNLP}, pages 1532--1543, Doha,
  Qatar.

\bibitem[\protect\citename{Poon and Domingos}2010]{poon-domingos:2010:ACL}
Hoifung Poon and Pedro Domingos.
\newblock 2010.
\newblock Unsupervised ontology induction from text.
\newblock In {\em Proceedings of the 48th Annual Meeting of the Association for
  Computational Linguistics}, pages 296--305, Uppsala.

\bibitem[\protect\citename{Rayner}1998]{rayner2009eye}
Keith Rayner.
\newblock 1998.
\newblock Eye movements in reading and information processing: 20 years of
  research.
\newblock {\em Psychological Bulletin}, 124(3):372--422.

\bibitem[\protect\citename{Rockt{\"a}schel \bgroup et al.\egroup
  }2016]{rocktaschel2015reasoning}
Tim Rockt{\"a}schel, Edward Grefenstette, Karl~Moritz Hermann, Tom{\'a}{\v{s}}
  Ko{\v{c}}isk{\`y}, and Phil Blunsom.
\newblock 2016.
\newblock Reasoning about entailment with neural attention.
\newblock In {\em Proceedings of the 2016 ICLR}, San Juan, Puerto Rico.

\bibitem[\protect\citename{Rush \bgroup et al.\egroup }2015]{rush2015neural}
Alexander~M Rush, Sumit Chopra, and Jason Weston.
\newblock 2015.
\newblock A neural attention model for abstractive sentence summarization.
\newblock In {\em Proceedings of the 2015 EMNLP}, pages 379--389, Lisbon,
  Portugal.

\bibitem[\protect\citename{Socher \bgroup et al.\egroup
  }2011]{socher2011dynamic}
Richard Socher, Eric~H Huang, Jeffrey Pennin, Christopher~D Manning, and
  Andrew~Y Ng.
\newblock 2011.
\newblock Dynamic pooling and unfolding recursive autoencoders for paraphrase
  detection.
\newblock In {\em Advances in Neural Information Processing Systems}, pages
  801--809.

\bibitem[\protect\citename{Socher \bgroup et al.\egroup
  }2013a]{socher-EtAl:2013:EMNLP}
Richard Socher, Alex Perelygin, Jean Wu, Jason Chuang, Christopher~D. Manning,
  Andrew Ng, and Christopher Potts.
\newblock 2013a.
\newblock Recursive deep models for semantic compositionality over a sentiment
  treebank.
\newblock In {\em Proceedings of the 2013 EMNLP}, pages 1631--1642, Seattle,
  Washington.

\bibitem[\protect\citename{Socher \bgroup et al.\egroup
  }2013b]{socher2013recursive}
Richard Socher, Alex Perelygin, Jean~Y Wu, Jason Chuang, Christopher~D Manning,
  Andrew~Y Ng, and Christopher Potts.
\newblock 2013b.
\newblock Recursive deep models for semantic compositionality over a sentiment
  treebank.
\newblock In {\em Proceedings of the 2013 EMNLP}, pages 1631--1642, Seattle,
  Washingtton.

\bibitem[\protect\citename{Sukhbaatar \bgroup et al.\egroup
  }2015]{sukhbaatar2015end}
Sainbayar Sukhbaatar, Jason Weston, Rob Fergus, et~al.
\newblock 2015.
\newblock End-to-end memory networks.
\newblock In {\em Advances in Neural Information Processing Systems}, pages
  2431--2439.

\bibitem[\protect\citename{Tai \bgroup et al.\egroup }2015]{tai2015improved}
Kai~Sheng Tai, Richard Socher, and Christopher~D Manning.
\newblock 2015.
\newblock Improved semantic representations from tree-structured long
  short-term memory networks.
\newblock In {\em Proceedings of the 53rd ACL}, pages 1556--1566, Beijing,
  China.

\bibitem[\protect\citename{Tanenhaus \bgroup et al.\egroup
  }1995]{Tanenhaus:ea:1995}
Michael~K. Tanenhaus, Michael~J. Spivey-Knowlton, Kathleen~M. Eberhard, and
  Julue~C. Sedivy.
\newblock 1995.
\newblock Integration of visual and linguistic information in spoken language
  comprehension.
\newblock {\em Science}, 268:1632--1634.

\bibitem[\protect\citename{Tran \bgroup et al.\egroup }2016]{ke2016memory}
Ke~Tran, Arianna Bisazza, and Christof Monz.
\newblock 2016.
\newblock Recurrent memory network for language modeling.
\newblock In {\em Proceedings of the 15th NAACL}, San Diego, CA.

\bibitem[\protect\citename{Wang and Jiang}2016]{wang2015learning}
Shuohang Wang and Jing Jiang.
\newblock 2016.
\newblock Learning natural language inference with lstm.
\newblock In {\em Proceedings of the 2016 NAACL: HLT}, pages 1442--1451, San
  Diego, California.

\bibitem[\protect\citename{Weston \bgroup et al.\egroup
  }2015]{weston2014memory}
Jason Weston, Sumit Chopra, and Antoine Bordes.
\newblock 2015.
\newblock Memory networks.
\newblock In {\em Proceedings of the 2015 ICLR}, San Diego, USA.

\bibitem[\protect\citename{Xiong \bgroup et al.\egroup }2016]{xiong2016dynamic}
Caiming Xiong, Stephen Merity, and Richard Socher.
\newblock 2016.
\newblock Dynamic memory networks for visual and textual question answering.
\newblock In {\em Proceedings of the 33rd ICML}, New York, NY.

\bibitem[\protect\citename{Yao \bgroup et al.\egroup }2015]{yao2015depth}
Kaisheng Yao, Trevor Cohn, Katerina Vylomova, Kevin Duh, and Chris Dyer.
\newblock 2015.
\newblock Depth-gated recurrent neural networks.
\newblock {\em arXiv preprint arXiv:1508.03790}.

\bibitem[\protect\citename{Zaremba and Sutskever}2014]{zaremba2014learning}
Wojciech Zaremba and Ilya Sutskever.
\newblock 2014.
\newblock Learning to execute.
\newblock {\em arXiv preprint arXiv:1410.4615}.

\end{thebibliography}
\end{document}